\pdfoutput=1

\documentclass[11pt]{article}

\usepackage[]{acl}

\usepackage{times}
\usepackage{latexsym}
\usepackage{tabularx}

\usepackage[T1]{fontenc}
\usepackage{array}
\usepackage{multirow}
\usepackage{graphicx}
\usepackage{subfigure}
\usepackage{array, diagbox}
\usepackage{wrapfig, blindtext, booktabs}
\usepackage{longtable}
\usepackage{enumitem}
\usepackage{caption}
\usepackage{amsmath}
\usepackage{xurl}
\usepackage[utf8]{inputenc}

\usepackage{microtype}

\newcommand{\stdintable}[1] {\textcolor{black}{\scriptsize{$\pm$#1}}}

\newcommand{\revised}[1]{\textcolor{black}{#1}}

\title{ODD: A Benchmark Dataset for the Natural Language Processing Based Opioid Related Aberrant Behavior Detection}

\author{Sunjae Kwon$^{1}$, Xun Wang$^{2}$, Weisong Liu$^{3}$, Emily Druhl$^{4}$, Minhee L. Sung$^{5}$, \\
  \textbf{Joel I. Reisman$^{4}$}, \textbf{Wenjun Li$^{3}$, Robert D. Kerns$^{5}$, William Becker$^{5}$, Hong Yu$^{1,3,4,6}$}\\
  $^{1}$UMass Amherst, $^{2}$Microsoft, $^{3}$UMass Lowell, $^{4}$U.S. Department of Veterans Affairs, \\$^{5}$Yale University,
  $^{6}$UMass Chan Medical School\\
  \texttt{\small sunjaekwon@umass.edu}, \texttt{\small wangxun.pku@gmail.com},\\
  \texttt{\small \{weisong\_liu, wenjun\_li, hong\_yu\}@uml.edu},
  \texttt{\small \{emily.druhl, joel.reisman\}@va.gov}, \\ \texttt{\small \{minhee.sung, robert.kerns, william.becker\}@yale.edu}}

\begin{document}
\maketitle

\begin{abstract}
Opioid related aberrant behaviors (ORABs) present novel risk factors for opioid overdose. This paper introduces a novel biomedical natural language processing benchmark dataset named ODD, for ORAB Detection Dataset. ODD is an expert-annotated dataset designed to identify ORABs from patients' EHR notes and classify them into nine categories; 1) Confirmed Aberrant Behavior, 2) Suggested Aberrant Behavior, 3) Opioids, 4) Indication, 5) Diagnosed opioid dependency, 6) Benzodiazepines, 7) Medication Changes, 8) Central Nervous System-related, and 9) Social Determinants of Health. We explored two state-of-the-art natural language processing  models (fine-tuning and prompt-tuning approaches) to identify ORAB. Experimental results show that the prompt-tuning models outperformed the fine-tuning models in most categories and the gains were especially higher among uncommon categories (Suggested Aberrant Behavior, Confirmed Aberrant Behaviors, Diagnosed Opioid Dependence, and Medication Change). Although the best model achieved the highest 88.17\% on macro average area under precision recall curve, uncommon classes still have a large room for performance improvement. ODD is publicly available\footnote{Dataset: \url{https://www.physionet.org/content/nlp-opioid-behavior-detection/1.0.0/}\\Source code: \url{https://github.com/soon91jae/ORAB_MIMIC}}.
\end{abstract}

\section{Introduction}

The opioid overdose (OOD) crisis has had a striking impact on the United States, not only threatening citizens' health \citep{azadfard2022opioid} but also bringing about a substantial financial burden \citep{florence2021economic}.
According to a report by the \citet{CDC2023}, OOD accounted for 110,236 deaths in a single year in 2022. 
In addition, fatal OOD and opioid use disorder (OUD) cost the United States \$1.04 trillion in 2017 and that figure rose sharply to \$1.5 trillion in 2021 \citep{Beyer2022}. Identifying patients at risk of OOD could help prevent serious consequences \citep{marks2021identifying}.

The opioid crisis is multifaceted, with factors like inadequate health insurance coverage \citep{blumenthal2017combat}, regulatory lapses \citep{kolodny2020fda}, and profit-motivated campaigns by pharmaceutical firms \citep{haffajee2017drug} contributing to its complexity. Countermeasures include deploying Prescription Drug Monitoring Programs (PDMPs) \citep{CDC2022}, enhancing addiction drug education for healthcare providers \citep{dowell2022cdc}, and developing less addictive drugs \citep{thomas2017amid}. Notably, PDMPs are data-driven systems tailored to detect patients at risk of OUD. By leveraging data analytics, these systems have successfully shielded many from critical OOD outcomes \citep{paulozzi2011prescription}. 

Opioid-Related Aberrant Behaviors (ORABs) or Aberrant Drug Related Behaviors (ADRBs) are patient behaviors that may indicate prescription medication abuse \citep{fleming2008reported}. ORABs can be categorized into confirmed aberrant behavior and suggested aberrant behavior \citep{portenoy1996opioid, laxmaiah2008monitoring, NIDA2023}. 
Herein, confirmed aberrant behaviors 
have a clear evidence of medication abuse and addiction while suggested aberrant behaviors do not have a clear evidence
\citep{NIDA2023}. Table~\ref{tab:example_of_ORABS} presents examples of such categories.

ORABs are not only clinically significant due to their strong association with OOD \citep{wang2022using} and drug misuse \citep{maumus2020aberrant}, but they also pose intriguing and challenging problems in terms of natural language processing (NLP). This is for two primary reasons. Firstly, unlike other BioNLP tasks where reliance is primarily on medical terms or jargon \citep{kwon-etal-2022-medjex}, ORABs encompass various behavioral patterns. These include attempts to deceive clinicians, contradictory statements, and scenarios that necessitate inference based on common sense. Secondly, given the rarity of ORABs in patients prescribed opioids \citep{nadeau2021opioids}, it's crucial to consider label bias.



\begin{table}[]
    \centering
    \footnotesize
    \resizebox{\linewidth}{!}{%
    \begin{tabular}{@{ }c@{}|@{ }l@{}}
    \hline
    ORAB Type & \multicolumn{1}{c@{}}{Example}\\
    \hline\hline
    \multirow{3}{*}{\begin{tabular}[c]{@{}l@{}}Confirmed\\Aberrant\\Behavior\end{tabular}} & Misuse of legal substances (e.g. Alcohol)                                                              \\
                             & Falsification of prescription—forgery or alteration\\
                             & Injecting medications meant for oral use \\
    \hline
    \multirow{3}{*}{\begin{tabular}[c]{@{}l@{}}Suggested\\Aberrant\\Behavior\end{tabular}} & Asking for or even demanding, more medication\\
                             & Asking for specific medications\\
                             &Reluctance to decrease opioid dosing once stable\\
    \hline
    \end{tabular}
    }
    \vspace{-2mm}
    \caption{ORAB examples}
    \vspace{-7mm}
    \label{tab:example_of_ORABS}
\end{table}

Previously, ORABs have been detected by monitoring opioid administration (e.g., frequency and dosage) \citep{rough2019using} or self-reported questionnaires \citep{adams2004development, webster2005predicting}. However such measurements do not include the full spectrum of ORABs (e.g., medication sharing, denying medication changing). In addition, patients can obtain opioids from multiple resources (e.g. illegal purchase and medication sharing), which are not captured in the structured data. It has been known that ORABs are widely described in EHR notes and NLP techniques can be used to identify ORABs \citep{lingeman2017detecting}. Nonetheless, the previous study relied on a small amount of annotated notes, which were not publicly available. Moreover, the previous work only considered ORABs as a binary classification (present or not) and only explored traditional machine learning models (e.g., support vector machine (SVM)).  

\revised{This paper proposes ORAB detection that is a novel Biomedical NLP (BioNLP) task. We also introduce an \textbf{O}RAB \textbf{D}etection \textbf{D}ataset \textbf{(ODD)} which is \textit{large-size}, \textit{expert-annotated}, and \textit{multi-label classification} benchmark dataset corresponding to the task.
For this, we first designed a robust and comprehensive annotation guideline that labels text into nine categories which encompass two types of ORABs (Confirmed Aberrant Behavior and Suggested Aberrant Behavior) and seven types of auxiliary opioid-related information (Opioids, Indication, Diagnosed Opioid Dependency, Benzodiazepines, Medication Change, Central Nervous System Related, Social Determinant of Health). Following the guideline, domain experts annotated 750 EHR notes from 500 opioid-treated patients in the MIMIC-IV database \citep{johnson2021mimic}, finding 399 notes with opioid prescriptions. In total, 3,718 instances were annotated across 2,940 sentences, including 162 instances of ORABs (115 Confirmed and 47 Suggested Aberrant Behavior instances).}

We conducted experiments on two Opioid-Related Aberrant Behavior (ORAB) detection models, employing state-of-the-art (SOTA) NLP models. These experiments utilized two distinct approaches: traditional fine-tuning, as described by \citet{devlin2018bert}, and prompt-based fine-tuning, following the methodology outlined by \citet{webson2022prompt}. The experimental results on MIMIC showed that prompt-based tuning models surpass fine-tuning models in almost all categories. \revised{Particularly noteworthy is the performance improvement in less common categories with fewer than 150 instances (referred to as $uncommon~categories$ such as Suggest Aberrant Behavior, Confirmed Aberrant Behaviors, Diagnosed Opioid Dependency, and Medication Change).} In these categories, the performance improvements were notably substantial, with the Diagnosed Opioid Dependency, Medication Change, and Suggested Aberrant Behavior classes each showing an increase of over 20 points.

The main contributions of this paper can be organized as follows:
\begin{itemize}
    \item This paper introduces a new BioNLP task \textbf{ORAB detection} for extracting information related to a patient's risk of opioid addiction and abuse from EHR notes. We also curate a corresponding benchmark dataset, named \textbf{ODD}, an expert-annotated dataset for the ORAB detection task. 
    \item We present the experimental results of two state-of-the-art NLP models as baseline performances for the benchmark dataset. Moreover, we report comprehensive data and error analyses to guide future studies in constructing improved models.
\end{itemize}

\section{Related Work}
\paragraph{NLP-based Opioid Abuse Analysis} Recently, with the development of NLP technology, studies have been
actively conducted to analyze information relevant to opioid abuse and OOD 
from text (e.g. EHR notes, social media) \citep{, sarker2019machine, blackley2020using, goodman2022development, zhu2022automatically, singleton2023using}. 
Studies have explored a broad range of NLP techniques to identify OUD 
\citep{zhu2022automatically}. \citet{zhu2022automatically} developed a keyword-based OUD detection model for patients who have been treated with chronic opioid therapy. 
Their NLP models were able to uncover OUD cases that would be missed using the International Classification of Diseases (ICD) codes alone. \citet{singleton2023using} proposed a multiple-phase OUD detection approach using a combination of dictionary and rule-based approaches. 
\citet{blackley2020using} developed feature engineering-based machine learning models. Herein, the authors demonstrated that the machine learning models outperformed a rule-based one that utilizes keywords. 

Other works adopted NLP to study factors associated with opioid abuse. \citet{goodman2022development} utilized text features such as term frequency–inverse document frequency (TF-IDF), concept unique identifier (CUI) embeddings, and word embeddings to analyze substances that contribute to opioid overdose deaths. 
\citet{sarker2019machine} conducted a geospatial and temporal analysis of opioid-related mentions in Twitter posts. They found a positive correlation between the rate of opioid abuse-indicating posts and opioid misuse rates and county-level overdose death rates. 

The ORAB detection task is similar to the studies above in that it analyzes drug abuse-related information using NLP approaches. However, different from the previous studies that mainly depend on keywords such as drug mentioning, the ORAB detection is a more challenging NLP task considering that it needs to identify various and complex linguistic patterns such as trying to deceive physicians \citep{passik2007commentary} and emotional reaction on opioid prescription \citep{lingeman2017detecting}.

\begin{table*}[!t]
\footnotesize
\centering
\renewcommand{\arraystretch}{1.2}
\resizebox{\linewidth}{!}{
\begin{tabular}{m{0.25\linewidth}|m{0.45\linewidth}|m{0.30\linewidth}}
\hline
\multicolumn{1}{@{}c@{ }|}{\textbf{Category}} & \multicolumn{1}{@{ }c@{ }|}{\textbf{Definition}} & \multicolumn{1}{@{ }c@{}}{\textbf{Example}} \\
\hline\hline
Confirmed Aberrant Behavior & Evidence confirming the loss of control of opioid use, specifically aberrant usage of opioid medications. & {[}Patient{]} admits that he has been sharing his Percocet with his wife, and that is why he has run out early. \\
\hline
Suggested Aberrant Behavior & Evidence suggesting loss of control of opioid use or compulsive/inappropriate use of opioids. & {[}Patient{]} states that ‘that {[}drug{]} won’t work; only {[}X drug{]} will and I won’t take any other’ \\
\hline
Opioids  & The mention or listing of the name(s) of the opioid medication(s) that the patient is currently prescribed or has just been newly prescribed. & Oxycodone has been known to make {[}the patient{]} sleepy at 5 mg. \\
\hline
Indication & Patients are using opioids under instructions. & {[}The patient{]} is in a daze. \\
\hline
Diagnosed Opioid Dependency & Patients have the condition of being dependent on opioids, have chronic opioid use, or is undergoing opioid titration & {[}The patient{]} is in severe pain and has been taking {[}opioid drug{]} for {[}time{]}.{[}HY1{]}               \\
\hline
Benzodiazepines & Patients are co-prescribed benzodiazepines. & Valium has been listed in patient medications. \\
\hline
Medicine Changes  & Change in opioid medicine, dosage, and prescription since the last visit. & [Patient] reports that his previous PCP just recently changed his pain regimen, adding oxycodone. \\
\hline
Central Nervous System Related & CNS-related terms/terms suggesting altered sensorium. & [Patient] reported to have nausea after taking {[}drug{]}. \\
\hline
Social Determinants of Health &  The nonmedical factors that influence health outcomes & [Patient] divorced a years ago.\\
\hline
\end{tabular}}
\vspace{-3mm}
\caption{The definitions and examples of the categories of ODD.}

\label{tab:label_examples}
\end{table*}

\paragraph{ORAB Risk Assessment and Detection} 
\citet{webster2005predicting} introduced a risk management tool that monitors ORABs by scoring a patient's self-reports on risk factors (history of family and personal substance abuse, history of preadolescent sexual abuse, and psychological illness) related to substance abuse. Then, each patient is categorized into three risk levels (low risk, moderate risk, and high risk) according to the sum of the scores. Other studies \citep{schloff2004identifying, sullivan2010risks, katz2010usefulness, tudor2013memorandum, rough2019using} suggest detecting ORAB by relying on diagnostic criteria based on structured information such as the frequency of opioid dosage, the number of opioid prescribers, and the number of pharmacies. Although the above methodologies can detect patients at risk of ORABs with high precision, the recall was low 
\citep{rough2019using}.

\citet{lingeman2017detecting} work is the most relevant to our study. 
However, as described earlier, \citet{lingeman2017detecting}'s work relied on a small scaled EHR notes which is not publicly available. In contrast, ODD consists of a larger dataset which is publicly available. Furthermore, ODD's annotation scheme provides rich sub-categorized aberrant behaviors (suggested and confirmed) and additional opioid-related information. 
In contrast, \citet{lingeman2017detecting}'s study was designed as a binary classification task to detect ORABs. Finally, we leverage the SOTA deep learning models that the previous work \citet{lingeman2017detecting} did not explore.

\section{Task Definition and Evaluation Criteria}
\paragraph{Task Definition} \revised{The ORAB detection is an \textbf{information extraction task} that identifies whether an input text contains ORABs (Confirmed, and Suggested aberrant behaviors) and additional concepts relevant to OOD and OUD. In addition, since all labels can be co-occurred together in a sentence, we formulate the \textbf{multi-label classification}.}

\paragraph{Evaluation Criteria} Previous study on NLP-based ORAB detection \citep{lingeman2017detecting} utilizes accuracy as an evaluation criterion. However, since the labels in the dataset are highly imbalanced (in Table~\ref{tab:annotation_statistics}), the accuracy may mislead performance on rare classes since it can overestimate true negative cases \citep{bekkar2013evaluation}.
\revised{Thus, as main evaluation criteria, we adopt the Area Under Precision-Recall Curve (AUPRC) and the F1-score that have been widely utilized for the performance evaluation of the binary classifiers on highly biased labels \citep{ozenne2015precision}.}

\section{ORAB Detection Dataset}
\subsection{Data Collection}

The source of the first dataset is made up of publicly available fully de-identified EHR notes of the MIMIC-IV \citep{Johnson2023-fg}. ORABs are uncommon events. To increase the likelihood that our annotated data incorporate ORABs, we sorted out patients at risk of opioid misuse based on repetitive opioid use and diagnosis related to opioid misuse. Specifically, we first extracted EHR notes mentioning opioids with the generic and brand name of opioid medications. In addition, we selected patients diagnosed based on their  ICD codes. Detailed information on opioid medications (and their generic names), and ICD codes utilized for filtering EHR notes are presented in Appendix~\ref{apx:data_cnstrct}.


Among 331,794 EHR notes of 299,712 patients in MIMIC-IV database, we found that approximately 57\% of patients were prescribed opioids during their hospitalization. Then, we selected patients who were repeatedly prescribed (more than twice) opioids.
In addition, we chose patients who were diagnosed with drug poisoning and drug dependence based on the ICD codes.
Overall, there are 3,904 patients who are satisfied the aforementioned conditions. Among them, we randomly select 750 notes from a randomly sampled 500 patients for annotation. 


\subsection{Data Annotation}

\begin{table}[]
\footnotesize
\centering
\begin{tabular}{c|c|r}
\hline
Socio-demographic type & Group & \# of patients (\%)\\
\hline
\hline
\multirow{2}{*}{Gender} & Male & 168 (51.69\%) \\
 & Female & 157 (48.31\%) \\
\hline
\multirow{7}{*}{Age} & 19-25 & 14 (4.31\%) \\
 & 26-35 & 34 (10.46\%) \\
 & 36-45 & 59 (18.15\%) \\
 & 46-55 & 80 (24.62\%) \\
 & 56-65 & 69 (21.23\%) \\
 & 66-75 & 40 (12.31\%) \\
 & $>$ 75 & 29 (8.92\%)\\
\hline
 \multicolumn{2}{c|}{Total}  & 325 (100\%)\\
\hline
\end{tabular}
\vspace{-2mm}
\caption{Socio-demographic statistics of the cohort. }
\vspace{-6mm}
\label{tab:demographic_statistics}
\end{table}

\begin{table}
\footnotesize
\centering
\begin{tabular}{l|r}
\hline
\multicolumn{1}{c|}{Categories} & \multicolumn{1}{c}{Instances} \\
\hline\hline
Confirmed Aberrant Behavior & 115 (3.09\%)\\
Suggested Aberrant Behavior &  47 (1.26\%)\\
Opioids                      & 1,678 (45.13\%)\\
Indication                  & 558 (15.01\%)\\
Diagnosed Opioid Dependency & 67 (1.80\%)\\
Benzodiazepines             & 417 (11.22\%)\\
Medication Change            & 139 (3.74\%)\\
Central Nervous System Related & 542 (14.58\%)\\
Social Determinants of Health & 155 (4.17\%)\\
\hline
Total                       & 3,718 (100\%)\\
\hline
\end{tabular}
\vspace{-2mm}
\caption{Categorical distribution of the annotated instances.}
\vspace{-5mm}
\label{tab:annotation_statistics}
\end{table}

\begin{figure*}
    \centering
    \begin{subfigure}
        \raggedleft
         \includegraphics[width=.39\linewidth]{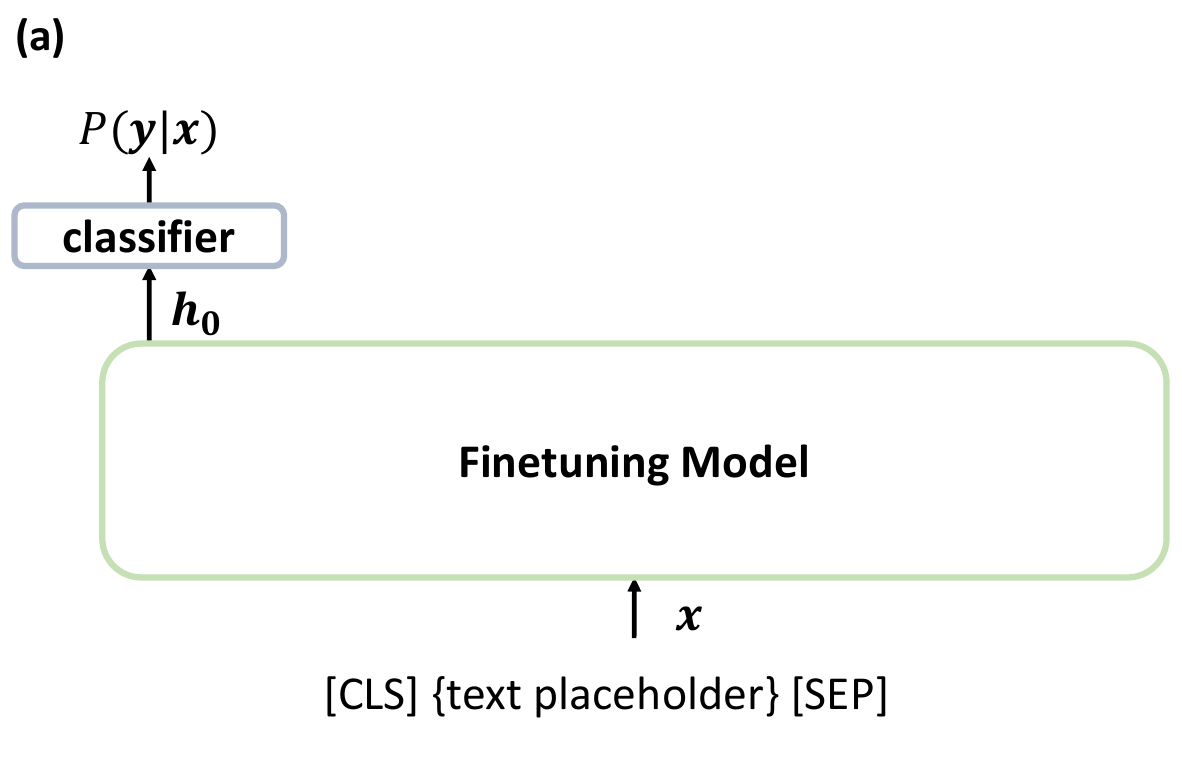}
         \label{fig:finetune}
     \end{subfigure}
     \begin{subfigure}
        \centering
            \includegraphics[width=.59\linewidth]{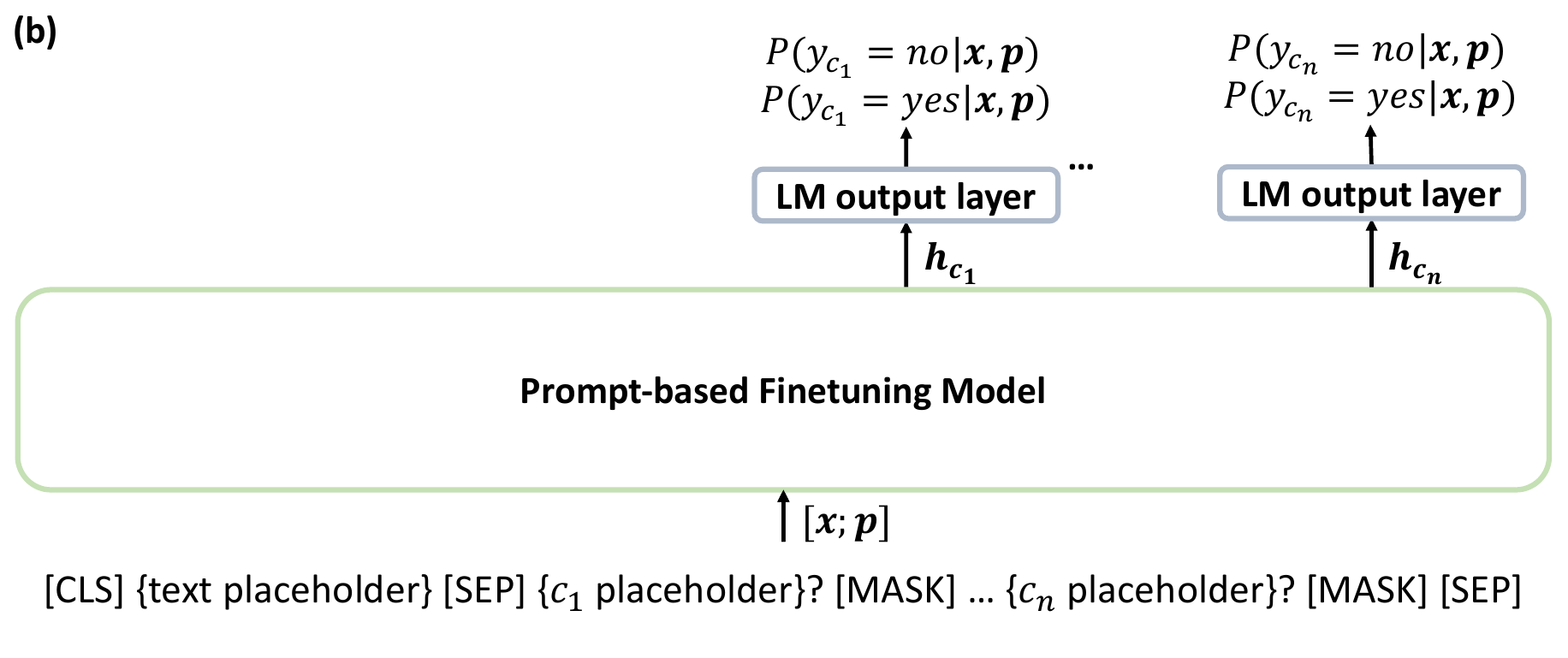}
         \label{fig:prompt}
     \end{subfigure}
    \caption{The figures illustrate the conceptual architectures of our ORAB detection models. (a) \revised{ demonstrates a fine-tuning model and (b) depicts a prompt-based fine-tuning model}. Herein, $\textbf{x}$, $\textbf{y}$, and $\textbf{p}$ indicate input text, output labels, and prompt text respectively. $\mathbf{h_i}$ is the hidden vector representation of the $i^{th}$ input token. EHR text input to `\{text placeholder\}'. The name of each category ($c_{1...n}$) in Table~\ref{tab:label_examples}  is input at `\{$c_{1...n}$ placeholder\}'.}
    \label{fig:benchmark_models}
\end{figure*}

\revised{For the annotation process, we initially identified nine clinically important categories for patients prescribed opioids who are either abusing them or at risk of opioid abuse. These categories include two types of ORABs – Confirmed Aberrant Behavior and Suggested Aberrant Behavior – as well as seven additional concepts. These concepts encompass opioid prescription (Opioids, and Indication) and risk factors associated with OUD or OOD (Diagnosed Opioid Dependency, Benzodiazepines, Medication Changes, Centeral Nervous System Related, Social Determinants of Health) \citep{darke1996fatal, jann2014benzodiazepines, arthur2018safe, mitra2021risk, kariisa2022vital} as briefly outlined in Table~\ref{tab:label_examples}.}

The annotation process was iterative, with continuous refinement of the EHR note annotations and annotation guidelines. An interdisciplinary team of addiction medicine, biostatisticians, and NLP specialists collaboratively discussed and developed these guidelines. This rigorous approach yielded a comprehensive annotation guideline adept at addressing language variations and ambiguities in clinical narratives related to opioid misuse. For detailed descriptions of the categories, please refer to Appendix~\ref{apx:annotation_scheme}. The annotation guidelines developed can be accessed in the 'annotation\_guideline.pdf' file available in the supplementary data.


EHR notes were annotated independently by two domain experts who are familiar with medical literature and EHR notes by following the annotation guidelines. Herein, the primary annotator \footnote{A master of public health} annotated all EHR notes with eHOST \citep{eHost} annotation tool. The other annotator \footnote{A medical doctor affiliated with the addiction medicine} coded 100 of the EHRs of the primary annotator with the same environment to compute inter-rater reliability with Cohen's kappa \citep{warrens2015five}. As a result, the inter-rater reliability shows strong agreement ($\kappa=0.86$) between the annotators. Detailed inter-rater reliability for each category can be found in Appendix~\ref{apx:IAA_per_category}. 

After annotation, among 750 notes, we could find 399 notes of 325 patients who are current opioid prescription. The socio-demographic statistics on the final patient cohort can be found in Table~\ref{tab:demographic_statistics}. Overall, there are 2,840 sentences that contain explicit evidences at least one of the target categories.

\subsection{Annotation Statistics}
Table 3 shows the statistics of the annotated instances from the 2,840 sentences. Herein, MIMIC dataset consist of 3,718 instances annotated from the EHRs. Especially, we can notice that `confirmed aberrant behavior' and `suggested aberrant behavior' in EHRs are relatively rare events only accounting for 162 (4.25\%); 115 (3.09\%) for confirmed aberrant behavior and 47 (1.26\%) for suggested aberrant behavior. The `Opioids,' `Indication,' and `Central nervous system related' are majority classes accounting for over 74\% of overall instances while the other categories are around or less than 10\% each. 


\section{ORAB Detection Models}

This section demonstrates pretrained Language Model (LM) based ORAB detections models; traditional fine-tuning model \citep{zahera2019fine} and prompt-tuning model. The prompt-based fine-tuning model has shown advantages in rare category classification (e.g. zero-shot or few-shot classification) \citep{yang2023multi}. Figure~\ref{fig:benchmark_models} demonstrates the baseline ORAB detection models.

\subsection{Fine-tuning Models}
\label{sec:fine_tuning}
The most common way to construct classification models using a pretrained language model (LM) is to employ fine-tuning, as illustrated in Figure~\ref{fig:benchmark_models}(a). In this approach, the input text $\mathbf{x}$ is passed through the fine-tuning model. The hidden representation vector of the first token `[CLS]' ($\mathbf{h_0}$) is then used as input for the classifier. Here, $W_c$ and $\mathbf{b_c}$ represent the weight matrix and bias, respectively. The classifier calculates the probability distribution over output labels $\mathbf{y}$ using the sigmoid function.


\subsection{Prompt-based Fine-tuning Models}
\label{sec:prompt-based_fine-tuning}

\begin{table*}[!ht]
\footnotesize
\centering
\renewcommand{\arraystretch}{1.2}
\resizebox{\linewidth}{!}{%
\begin{tabular}{@{}c@{ }|@{ }l@{ }l@{ }|@{ }l@{ }l@{ }|@{ }l@{ }l@{ }|@{ }l@{ }l@{}}
\hline
\multirow{3}{*}{Categories} & \multicolumn{4}{@{ }c@{ }|@{ }}{Fine-tuning} & \multicolumn{4}{@{ }c@{ }}{Prompt-based Fine-tuning} \\
\cline{2-9}
 & \multicolumn{2}{@{ }c@{ }|@{ }}{BioBERT} & \multicolumn{2}{@{ }c@{ }|@{ }}{BioClinicalBERT} & \multicolumn{2}{@{ }c@{ }|@{ }}{BioBERT} & \multicolumn{2}{@{ }c@{ }}{BioClinicalBERT} \\
\cline{2-9}
 & \multicolumn{1}{@{ }c@{ }@{ }}{AUPRC} & \multicolumn{1}{@{ }c@{ }|@{ }}{F1} & \multicolumn{1}{@{ }c@{ }@{ }}{AUPRC} & \multicolumn{1}{@{ }c@{ }|@{ }}{F1} & \multicolumn{1}{@{ }c@{ }@{ }}{AUPRC} & \multicolumn{1}{@{ }c@{ }|@{ }}{F1} & \multicolumn{1}{@{ }c@{ }@{ }}{AUPRC} & \multicolumn{1}{@{ }c@{ }@{ }}{F1} \\
\hline
\hline
Confirmed Aberrant Behaviors&77.79\stdintable{4.80}&64.07\stdintable{6.68}&79.48\stdintable{2.53}&66.19\stdintable{5.09}&\textbf{84.46\stdintable{16.40}}&\textbf{71.44\stdintable{12.02}}&\textbf{90.52\stdintable{6.54}}&\textbf{78.25\stdintable{7.07}}\\
Suggested Aberrant Behaviors&25.80\stdintable{4.10} & 23.62\stdintable{4.18} & 25.83\stdintable{2.36} & 29.39\stdintable{5.16} & \textbf{46.07\stdintable{17.87}} & \textbf{46.93\stdintable{12.77}} & \textbf{46.04\stdintable{16.27}} & \textbf{44.37\stdintable{13.02}}\\
Opioids&98.91\stdintable{0.30} & 97.29\stdintable{0.59} & 99.04\stdintable{0.23} & 97.23\stdintable{0.41} & \textbf{99.55\stdintable{0.16}} & \textbf{97.52\stdintable{0.47}} & \textbf{99.57\stdintable{0.18}} & \textbf{98.00\stdintable{0.29}}\\
Indication&97.57\stdintable{0.77} & \textbf{94.77\stdintable{0.45}} & 97.29\stdintable{1.02} & \textbf{94.25\stdintable{1.11}} & \textbf{97.83\stdintable{0.90}} & 93.89\stdintable{1.60} & \textbf{97.86\stdintable{0.90}} & 93.55\stdintable{0.94}\\
Diagnosed Opioid Dependency&60.54\stdintable{6.96} & 45.04\stdintable{4.86} & 61.54\stdintable{4.49} & 49.63\stdintable{4.45}	& \textbf{88.67\stdintable{10.83}} & \textbf{80.90\stdintable{10.09}} & \textbf{90.15\stdintable{7.22}} &\textbf{79.24\stdintable{12.97}} \\
Benzodiazepines&96.83\stdintable{1.01} & 95.09\stdintable{1.27} & 96.47\stdintable{1.33} & 94.40\stdintable{0.94} & \textbf{97.39\stdintable{1.33}} & \textbf{95.43\stdintable{0.82}} &\textbf{96.89\stdintable{1.71}} & \textbf{97.15\stdintable{1.55}} \\
Medication Change&51.64\stdintable{4.13} & 46.42\stdintable{2.12} & 56.02\stdintable{5.52}&50.72\stdintable{1.65}& \textbf{79.21\stdintable{3.46}} & \textbf{68.32\stdintable{1.67}}& \textbf{76.33\stdintable{4.06}} & \textbf{68.61\stdintable{5.14}} \\
Central Nervous System Related&97.83\stdintable{0.57} & 87.10\stdintable{2.10}& 98.15\stdintable{0.46} &88.11\stdintable{1.00}& \textbf{98.60\stdintable{1.23}}& \textbf{94.85\stdintable{1.25}} & \textbf{98.74\stdintable{0.53}}& \textbf{92.81\stdintable{2.44}} \\
Social Determinants of Health&94.32\stdintable{1.07}&80.20\stdintable{5.73}&92.82\stdintable{1.81}&83.90\stdintable{6.30}&\textbf{96.17\stdintable{2.01}}&\textbf{93.65\stdintable{4.04}}& \textbf{97.39\stdintable{1.83}} & \textbf{93.79\stdintable{3.08}} \\
\hline
Macro Average&77.91\stdintable{26.36}&70.40\stdintable{26.81}&78.52\stdintable{25.63}&72.65\stdintable{24.58}& \textbf{87.55\stdintable{17.12}}&\textbf{82.55\stdintable{17.29}}&\textbf{88.17\stdintable{17.40}} & \textbf{82.86\stdintable{17.62}}\\
\hline

\end{tabular}}
\caption{This table presents the experimental results of ODD on BioClinicalBERT and BioBERT. Each value stands for the average and the standard deviation of test folds of the nested cross-validation results and average scores with higher values between Fine-tuning and Prompt-based Fine-tuning marked as bold.}
\label{tab:experimental_results}
\end{table*}

\revised{While fine-tuning pre-trained LMs has been widely successful in various NLP tasks \citep{devlin2018bert}, it often requires a substantial number of annotated examples to achieve high performance \citep{webson2022prompt, yang2023multi}. This requirement can be particularly challenging for categories in Opioid Dependency Detection (ODD) that have fewer instances, potentially becoming a bottleneck in performance. Prompt-based fine-tuning, a technique highlighted in the works of \citet{gao2021making} and \citet{yang2022knowledge}, addresses this issue. It involves fine-tuning models using a template to reframe a downstream task as a language modeling problem. This is achieved by integrating masked language modeling with a pre-defined set of label words. Prompt-based fine-tuning is especially effective in few-shot scenarios, where the available training data is limited, and is known to outperform traditional fine-tuning methods in such contexts.}

We utilize the full name of each class to curate the prompt text $p$. Specifically, the prompts for each class are arranged in the same order as Table~\ref{tab:example_of_ORABS}, following the template ``\{$c_i$ placeholder\}? [MASK]'' where $c_i$ represents the name of the $i^{th}$ class. The prompt text is then concatenated with $\mathbf{x}$, distinguished by a separator token ``[SEP],'' and fed into a prompt-based tuning model. Next, we calculate the probability that the language model (LM) output of the masked token corresponding to each class would be a positive word or a negative word. Following the approach of \citet{gao2021making}, we define the positive word as `yes' and the negative word as `no'. 
Thus, the probability of `yes' for the $i^{th}$ class $c_i$ 
($P(y_{c_i}=$`yes'$|\mathbf{x},\mathbf{p})$) 
can be interpreted as the probability that $c_i$ is included in the input text $\mathbf{x}$, and vice versa.

\section{Experiment}
\subsection{Experimental Environment}
\paragraph{Experimental Models} \revised{To verify the generalizability of experimental results, we utilized two different LMs pretrained on Biomedical literature; BioBERT \citep{lee2020biobert} and BioClinicalBERT \citep{alsentzer2019publicly}.}  

\paragraph{Experimental Setting}
\revised{For the experiments, we conducted the nested cross-validation \citep{muller2016introduction} where outer and inner loops are 5 and 2 respectively. We choose the hyper-parameters for each outer loop that achieved the best performance on the inner folds with the grid search with the following range of possible values for each hyper-parameter: \{2e-5, 3e-5, 5e-5\} for learning rate, \{4, 8, 16\} for batch size, \{3,4,5\} for the number of epoch. Then, we report the average performance and standard deviation.}

To evaluation the significance of the performance differences between two models, we conducted Wilcoxon signed-rank test \citep{woolson2007wilcoxon}. In all of the experiments, we keep the random seed as 0. Finally, all experiments were performed on an NVIDIA P40 GPU with CentOS 7 version.

\subsection{Experimental Results}

\revised{Table~\ref{tab:experimental_results} displays the experimental results. Models achieved a performance range of [77.91, 88.17] in macro average AUPRC and [70.40, 82.86] in macro average F1.} Notably, the prompt-based fine-tuning models significantly outperformed the standard fine-tuning models in both the BioClinicalBERT and BioBERT frameworks, with an increase of 9.64 points and 9.65 points in macro AUPRC, respectively. Models based on BioClinicalBERT showed higher macro average performance than those based on BioBERT. This aligns with expectations, considering that BioClinicalBERT was pre-trained on EHR notes from MIMIC-III \citep{johnson2016mimic}, the predecessor to our target MIMIC-IV database, with both datasets sourced from the same hospital.

 
The performance disparity across different classes is notable. For instance, in the BioClinicalBERT fine-tuning model, the AUPRC score for the highest-performing class, Opioids, is 99.04, which is more than triple the score of the lowest-performing class, Suggested Aberrant Behaviors, at 25.83. This performance gap correlates with the number of instances in each class. Dominant classes like Opioids, Indication, Benzodiazepines, and Central Nervous System Related exhibit high performance with scores of 99.04, 97.29, 96.47, and 98.15, respectively. In contrast, less common categories show lower performance, with Suggested Aberrant Behavior at 25.83, Confirmed Aberrant Behavior at 79.48, Diagnosed Opioid Dependency at 61.54, and Medication Change at 56.02. Similar trends in performance are observed in the BioBERT model.



\revised{Prompt-based fine-tuning contributes to significantly enhanced macro average scores both BioBERT and BioClinicalBERT ($p<.01$). In all cases, prompt-based fine-tuning shows higher performance than fine-tuning, except for the F1 score of the indication class where is negligible (-0.88 points). The introduction of prompt-based fine-tuning resulted in significant improvements ($p<.01$), particularly in uncommon categories. The AUPRC of prompt-based fine-tuning on BioClinicalBERT and BioBERT increased by 20.21 and 20.27 points respectively in the Suggested Aberrant Behavior. In the Diagnosed Opioid Dependence, the AUPRC of prompt-based fine-tuning on BioClinicalBERT and BioBERT improved by 28.61 points and 28.13 points, respectively. Lastly, in the Medication Change class, the performance saw a rise of more than 20 points on BioBERT. From these results, we can infer that the low performance of uncommon categories is related to the sparsity of the instance, and that it can be improved through methods that enable effective learning with small data, such as prompt-based fine-tuning. However, we can also see that further performance improvements are still needed for uncommon categories.}


\begin{figure}
    \centering
    \includegraphics[width=\linewidth]{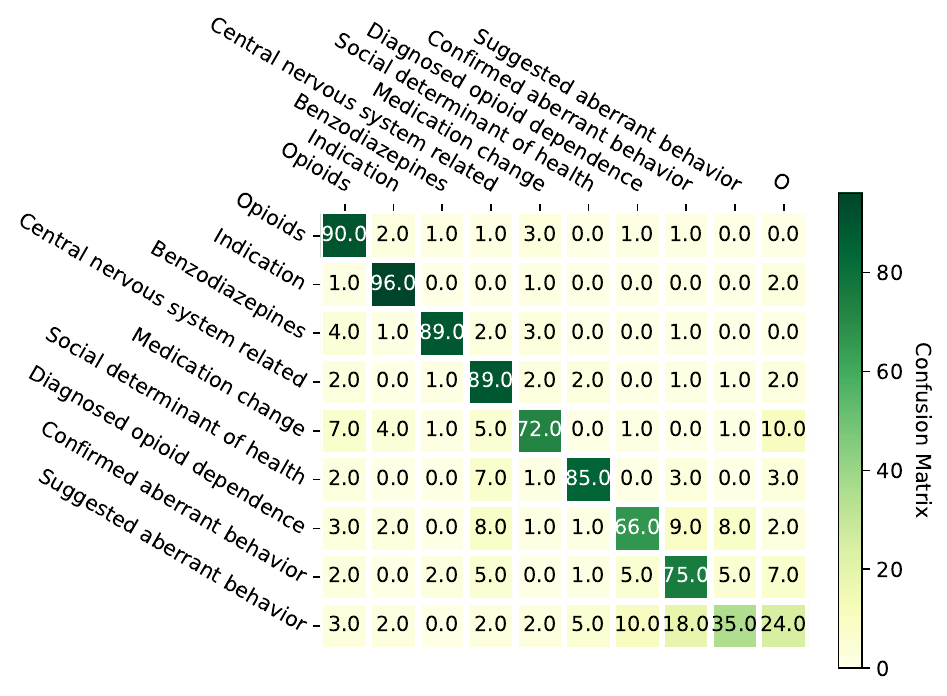}
    \vspace{-7mm}
    \caption{A multi-label confusion matrix among categories. `O' indicates the none of any categories.}
    \label{fig:confusion_matrix}
    \vspace{-3mm}
\end{figure}

\section{Discussion}

\subsection{Error Analysis}

\revised{First of all, we demonstrate quantitative aspects of errors.} For this, we gathered all the results of the test sets of 5-fold cross validation then calculated a normalized multi-label confusion matrix \citep{heydarian2022mlcm}. Figure~\ref{fig:confusion_matrix} shows that there are confusions between two specific classes: confirmed aberrant behavior and suggested aberrant behavior. The confusion rates were found to be 5.0\% and 18.0\%, respectively, for these classes. This indicates that the confirmed and suggested aberrant behaviors were the classes most prone to being mistaken for one another in our test sets. In addition, there are large confusions among diagnosed opioid dependence, confirmed and suggested aberrant behaviors which is 17\% in total.

\begin{table}[]
\centering
\footnotesize
\renewcommand{\arraystretch}{1.1}
\begin{tabular}{l|c}
\toprule
\multicolumn{2}{c}{Suggested Aberrant Behaviors} \\
\hline\hline
Confused as Confirmed Aberrant Behaviors     & 2  \\
Clinician's concern about non-opioid        & 1     \\
Annotation error        & 1     \\
Patient’s request for a higher or specific opioid & 1 \\
Obtaining opioids from multiple-medical sources & 2\\
Obtaining opioids from non-medical sources & 1\\
Others                                       & 2  \\
\midrule
\multicolumn{2}{c}{Confirmed Aberrant Behaviors}     \\
\hline\hline
Opioid use without evidence of abusing       & 2     \\
History of substance abuse                   & 3     \\
Confusion as Suggested Aberrant Behaviors  & 2     \\
Negation                                     & 2     \\
Substance abuse OTHER than prescription opioids & 1  \\
Self-escalating dose & 3 \\

\midrule
\multicolumn{2}{c}{Medication Change}                \\
\hline\hline
Previous medication change                   & 7     \\
Recommendation                               & 3     \\
Clinician's refusal to change medication     & 3     \\
Follow-up appointment for medication changes & 1     \\
Annotation error                             & 2     \\
Others                                       & 1    \\
\midrule
\multicolumn{2}{c}{Diagnosed Opioid Dependence}     \\
\hline\hline
Substance use disorder other than opioids     & 1 \\
Suspection                           & 2     \\
Other     & 1     \\
\bottomrule
\end{tabular}
\caption{Error analysis results on sampled data}
\label{tab:error_cases}
\end{table}

\revised{We conducted a qualitative error analysis on 100 cases from the predictions of the BioClinicalBERT prompt-based fine-tuning model. Table~\ref{tab:error_cases} presents these results, focusing on categories with F1 scores below 80\%: Suggested Aberrant Behavior, Confirmed Aberrant Behaviors, Diagnosed Opioid Dependency, and Medication Change.}

\revised{10 errors were found in the case of Suggested Aberrant Behaviors. In particular, confusion with Confirmed Aberrant Behavior and failure to predict obtaining opioids from multiple medical sources were the most representative errors. In addition, concerns about non-opioids (e.g. suicide) are representative error cases.}

\revised{Confirmed Aberrant Behaviors were found 13 times. Two of these were confusion with Suggested Aberrant Behavior, such as 'doctor shopping'. Meanwhile, `negation' cases denying opioid abuse, such as ``no pain med seeking behavior.'', were also found twice, and there were also 2 errors related to recording substance abuse. There is one case where it was not detected even when there was evidence of obvious substance abuse, such as alcoholism.}

\revised{Medication change errors accounted for the largest proportion of the cases, at 16. Among them, the most cases are records of previous medication changes. In addition, three errors occurred in each case related to recommendation or refusal of medication change. In addition, we could find one mention of an appointment to discuss future medication changes and two annotation errors.}

\revised{Finally, diagnosed opioid dependence refers to dependence on substances other than opioids, such as ``ETOH \footnote{Ethanol} dependence'', which is related to Confirmed Aberrant Behaviors. In addition, two cases in which opioid dependence or abuse were suspected but no apparent diagnose were specified.}

\subsection{Socio-demographic Analysis}

\begin{table}[]
    \centering
    \footnotesize
    \renewcommand{\arraystretch}{1.1}
    \resizebox{\columnwidth}{!}{
    \begin{tabular}{@{}c@{ }|@{ }c@{ }c@{ }|@{ }c@{ }c@{}} 
         \hline
        \multirow{3}{*}{} & \multicolumn{2}{@{ }c@{ }|@{ }}{Age} & \multicolumn{2}{c}{Gender}\\ 
        \cline{2-5} 
        & $<$45 & $\ge$ 45 & Female & Male\\ 
        \cline{2-5}
         & AUPRC & AUPRC & AUPRC & AUPRC \\ 
        \hline
        \hline
        CAB &\textbf{95.84\stdintable{4.95}}&88.36\stdintable{8.03}&89.29\stdintable{8.46}&89.15\stdintable{8.55}\\ 
        SAB &\textbf{60.77\stdintable{14.09}}&36.68\stdintable{17.79}&51.34\stdintable{14.20}&47.99\stdintable{20.16}\\ 
        \hline
    \end{tabular}
    }
    \caption{Experimental results on different age and gender groups. CAB and SAB mean   confirmed aberrant behaviors and suggested aberrant behaviors, respectively.}
    \label{tab:disaggregation_experiments}
\end{table}


Patient groups with varying socio-demographics frequently exhibit distinct characteristics. To examine the disparities among these groups, we carried out disaggregate studies that on two socio-demographic factors (age and gender) in Table~\ref{tab:disaggregation_experiments}.

\paragraph{Gender} The gender of the patients has little effect on the aberrant behavior detection performance, which means that the bias between genders is trivial. In fact, the male and female groups account for almost the same proportion of the total number of patients.

\begin{table}[]
\centering
\footnotesize
\renewcommand{\arraystretch}{1.1}
\resizebox{\linewidth}{!}{
\begin{tabular}{@{}l@{ }|@{ }c@{ }@{ }c@{}}
\hline
\multicolumn{1}{c@{ }|@{ }}{Confirmed Aberrant Behaviors}  & \multicolumn{2}{c}{Age}  \\
\hline
\multicolumn{1}{c@{ }|@{ }}{Subcategories} & < 45 & $\ge$ 45 \\
\hline\hline
Self-escalating dose & 1 & 5 \\
Using opioids outside of the prescriber's purpose & 1 & 3 \\
Substance abuse OTHER than prescription opioids & 0 & 2 \\
Evidence of a patient selling or giving opioids to others & 0 & 1 \\
\hline\hline
\multicolumn{1}{c@{ }|@{ }}{Suggested Aberrant Behaviors} & \multicolumn{2}{c}{Age} \\
\hline
\multicolumn{1}{c@{ }|@{ }}{Subcategories} & < 45 & $\ge$ 45 \\
\hline\hline
Clinician's concern on opioids & 2 & 1 \\
Obtaining opioids from non-medical sources & 0 & 2 \\
Patient's request for a higher or specific opioid & 3 & 3 \\
Obtaining opioids from multiple-medical sources& 1 & 2 \\
Patient's strong emotion/opinion on opiods & 0 & 1 \\
Others & 1 & 1\\
\hline
\end{tabular}}
\vspace{-2mm}
\caption{Subcategorical error analysis on different age groups.}
\vspace{-3mm}
\label{tab:subcategorical_analysis}
\end{table}

\paragraph{Age} We divided patients into two groups based on age 45, which is the standard for specifying the risk according to the patient's age \citep{brott2020opioid}, and evaluated performance of aberrant behaviors. Experimental results showed that the performances of aberrant behaviors are significantly different between two age groups. Especially, the performance of the younger age group achieved higher performance although the proportion of patients in the older group is greater (over 45: 69.23\%, less than 45: 30.77\%).

We speculate that this is because more diverse patterns of aberrant behaviors are observed in the older group. Table~\ref{tab:subcategorical_analysis} shows the error analysis results for each age group. We can see that both confirmed aberrant behaviors and suggested aberrant behaviors in the older group show more diverse aberrant behavior patterns than in the younger group.

\subsection{\revised{Prospective Social Impact of the Dataset}}
Our research can have the following positive impacts. Firstly, the information extracted by ORAB detection models can be utilized for various studies and systems aimed at addressing opioid abuse. For instance, since ORABs serve as important evidence of OUD, they can be used as key features in opioid risk monitoring systems. Additionally, this information can be leveraged to detect a patient's risk of OOD or opioid addiction at an earlier stage, thereby assisting in the prevention of fatal OOD cases. Consequently, by supporting efforts to mitigate future opioid overdoses, our research would contribute to maintaining people's health.

However, it is important to acknowledge that our work may have certain negative social impacts. As previously mentioned, ORAB detection can be utilized to strengthen opioid monitoring systems, but this may unintentionally encroach upon the autonomy of doctors \citep{clark2012prescription}. Indeed, in previous studies, although strict opioid prescription policies and prescription PDMPs help patients forestall opioid misuse or overuse \citep{mccauley2016dental, dowell2016mandatory}, oligonalgesia \citep{dowell2016mandatory}, has been pointed out as a possible side effect of PDMPs \citep{cantrill2012clinical}.

\section{Conclusion}
This paper introduces a novel BioNLP task called ORAB detection, which aims to identify two ORAB categories and seven categories relevant to opioid usage from EHR notes. We also present the associated benchmark dataset, ODD. The paper provides baseline models and their performances on ODD. To this end, we trained two SOTA pretrained LMs using a fine-tuning approach and prompt-based fine-tuning. Experimental results demonstrate that the performance in three uncommon categories was notably lower compared to the other categories. However, we also discovered that prompt-based fine-tuning can help mitigate this issue. Additionally, we provide various error analysis results to guide future studies.

\section*{Ethical Consideration}
First, one prospective concern is whether is it legal to screen patients and provide prior medical history without their consent. According to the \citet{HHS2021}, “The Health Insurance Portability and Accountability Act (HIPAA) regulation allows health care providers to disclose protected health information about an individual, without the individual’s authorization, to another health care provider for that provider’s treatment of the individual” (§ 45 CFR 164.506 ). Health care providers can be defined at §45 CFR PART 171 \citep{HIT2020}:

\begin{itemize}
    \item hospital, skilled nursing facility, nursing facility, home health entity or other long-term care facility, health care clinic, community mental health center, renal dialysis facility, blood center, ambulatory surgical center, emergency medical services provider, Federally qualified health center, group practice, a pharmacist, a pharmacy, a laboratory, a physician, a practitioner, a provider operated by, or under contract with, the Indian Health Service or by an Indian tribe, tribal organization, or urban Indian organization, a rural health clinic, a covered entity under section 256b of this title, an ambulatory surgical center, a therapist, and any other category of health care facility, entity, practitioner, or clinician determined appropriate by the Secretary.
\end{itemize}

Another consideration is the dataset's quality. We attempted to ameliorate this issue by developing a thoroughly systematic annotation guideline. First of all, we used an iterative process throughout the annotation, going back and forth between EHR note annotations and establishing annotation guidelines. The guidelines were discussed among an interdisciplinary team of experts in addiction (3), biostatisticians (2), and NLP (2). In this process, we curated a comprehensive annotation guideline, which addresses various aspects of how to handle language variations and ambiguities in clinical narratives related to this annotation task.

In addition, the data annotation quality might be a concerned since it requires specialized medical knowledge. Although the main annotator's annotations are almost perfectly aligned with the domain expert ($\kappa=0.86$), it is still a question whether the primary annotator is consistent. Thus, to analyze annotation quality, the primary annotator performed re-annotation on 25 sampled notes. At this time, initial annotation was performed on April 21-May 26, and re-annotation was performed on August 25-26, about 3 months later. The Kappa score of the two annotations was $\kappa=0.96$, which was almost perfectly consistent with the previous annotations. This implies that the annotation of the dataset used in this paper is consistent and reliable.

All EHR data we used in this paper were obtained through legal channels. Authors and annotators acquired eligible licenses to change and publish data. All data annotators are full-time employees.

\section*{Limitation \& Future Work}
The ORAB detection task relies on EHR notes. Thus, if health providers do not recognize the patient's abnormal signs, they may not describe aberrant behaviors in a note. In this case, our approach cannot detect ORABs. In the future, we will develop an algorithm that detects a wider spectrum of ORABs by combining them with previous structured information-based methods.

\revised{Another limitation is that our data source was derived from a single hospital's EHR database. Although many existing studies have been conducted based on the MIMIC database, this does not guarantee that the system developed as a result of this study can be migrated to different clinical settings. In addition, the dataset targets only single language English that is a limiation in analyzing EHR notes written in various languages. Thus, it is required to perform annotation based on annotation guidelines in additional clinical environments. Moreover, some categories such as `Social Determinant of Health' defined too broad. Thus, it is required to adopt fined-grained Social Determinant of Health extraction models \citep{ahsan2021mimic} to use in a real environment. }

ORAB detection models still have limited performance in the uncommon categories. \revised{It is necessary to improve performance through advanced NLP approaches such as data augmentation \citep{wei2019eda}, medical knowledge injection \citep{yang2022knowledge}, or leveraging knowledge extracted from generative Large Language Models (LLMs) \citep{kwon2023vision}.} \revised{Indeed, one prospective application of utilizing generative LLMs on this task is data augmentation. For example, we additionally conducted data augmentation experiments with a LLM, Flan T5 XL \citep{chung2022scaling}, for data augmentation with a simple prompt.}

\begin{center}
    ``Rewrite: \{input text holder\}''
\end{center}

\revised{Here, we generated three paraphrased sentences for all sentences of the train set of each fold and add them to the training set. Experimental results showed that the data augmentation helps to enhance the performance of aberrant behavior detection at BioClinicalBERT + Prompt-based environment.}

\revised{The results in Table~\ref{tab:paraphrasing} demonstrate that data augmentation with generative LLMs could be a promising solution for this task achieving higher performance. However, due to the various linguistic patterns of suggested aberrant behaviors, there is still room for performance improvement by paraphrasing alone. Through developed data augmentation method with LLMs in the future, we can expect additional performance improvements in suggested aberrant behaviors and medication change classes. Entire experimental results containing additional categories can be found in Appendix~\ref{apx:data_aug}.}

\begin{table}[]
\centering
\footnotesize

\begin{tabular}{@{}c@{ }|@{ }c@{ }c@{ }|@{ }c@{ }c@{}}
\hline
\multicolumn{1}{@{}c@{ }|@{ }}{\multirow{2}{*}{}} & \multicolumn{2}{@{}c@{ }|@{ }}{BioClinicalBERT} & \multicolumn{2}{c}{T5 Paraphrasing} \\
\cline{2-5}
\multicolumn{1}{c@{ }|@{ }}{} & AUPRC & F1 & AUPRC & F1 \\
\hline
CAB & 90.52\stdintable{6.54} & 78.25\stdintable{7.07} & \textbf{93.86\stdintable{4.53}} & \textbf{87.36\stdintable{6.41}} \\
SAB & 46.04\stdintable{16.27} & 49.70\stdintable{13.02} & \textbf{65.63\stdintable{16.02}} & \textbf{57.30\stdintable{14.65}}\\
\hline
\end{tabular}
\caption{Experimental results of the data augmentation with the LLM paraphrasing on confirmed aberrant behaviors (CAB) and suggested aberrant behaviors (SAB).}
\vspace{-4mm}
\label{tab:paraphrasing}
\end{table}

\revised{Finally, errors can cause negative downstream effects. In particular, the most significant negative downstream impact is that some errors for example misprediction of opioid dependences or ORABs can lead to a false stigma to the patient which is known as one of the unintended harms of PDMPs reducing the quality of medical care \citep{haines2022patient}. A way to alleviate this problem is to not only provide predictions to clinicians, but also provide rationale for them so that they can judge the patient's condition from various perspectives \citep{walsh2020stigma}. However, it is difficult to expect providing rationales for prediction with our approaches that utilized pre-trained LM-based extraction models since interpretable output cannot be generated due to the structure of the models.}

\section*{Acknowledgement}
Research reported in this publication was supported in part by the National Institutes of Health (NIH) under award numbers R01DA056470 and NIH R01DA045816. This work was also supported in part by 1I01HX003711-01A1 from the United States (U.S.) Department of Veterans Affairs Health Systems Research (HSR). The content is solely the responsibility of the authors and does not necessarily represent the official views of the NIH and HSR.

\bibliography{anthology,custom}
\bibliographystyle{acl_natbib}

\newpage
\onecolumn
\appendix

\section{Details on Data Construction}
\label{apx:data_cnstrct}
\subsection{Details on Data Collection}

\begin{center}
\footnotesize
\begin{longtable}{l|l}
\caption{Opioids and their generic naming that used for filtering.}\\

\hline \multicolumn{1}{c}{Medication Names} & \multicolumn{1}{|c}{Generic Names} \\ \hline 
\endfirsthead

\multicolumn{2}{c}%
{{ \tablename\ \thetable{}: continued from previous page}} \\
\hline \multicolumn{1}{c}{Medication Names}  & \multicolumn{1}{|c}{Generic Names}\\ \hline 
\endhead

\hline \multicolumn{2}{r}{{Continued on next page}} \\
\endfoot

\hline\hline
Ascomp with Codeine                 & aspirin/butalbital/caffeine/codeine       \\
B \& O Supprettes                   & belladonna/opium                          \\
Darvon Compound-65                  & aspirin/caffeine/propoxyphene             \\
Lorcet                              & acetaminophen/hydrocodone                 \\
Maxidone                            & acetaminophen/hydrocodone                 \\
Fiorinal with Codeine III           & aspirin/butalbital/caffeine/codeine       \\
Magnacet                            & acetaminophen/oxycodone                   \\
Meprozine                           & meperidine/promethazine                   \\
Fiorinal with Codeine               & aspirin/butalbital/caffeine/codeine       \\
Fioricet with Codeine               & acetaminophen/butalbital/caffeine/codeine \\
Lorcet Plus                         & acetaminophen/hydrocodone                 \\
Percocet 10 / 325                   & acetaminophen/oxycodone                   \\
Primlev                             & acetaminophen/oxycodone                   \\
Suboxone                            & buprenorphine/naloxone                    \\
Ibudone                             & hydrocodone/ibuprofen                     \\
Lorcet 10 / 650                     & acetaminophen/hydrocodone                 \\
Panlor DC                           & acetaminophen/caffeine/dihydrocodeine     \\
Reprexain                           & hydrocodone/ibuprofen                     \\
Percocet                            & acetaminophen/oxycodone                   \\
Combunox                            & ibuprofen/oxycodone                       \\
Hydrocet                            & acetaminophen/hydrocodone                 \\
Roxicet                             & acetaminophen/oxycodone                   \\
Tylox                               & acetaminophen/oxycodone                   \\
Xolox                               & acetaminophen/oxycodone                   \\
Vicodin ES                          & acetaminophen/hydrocodone                 \\
Hycet                               & acetaminophen/hydrocodone                 \\
Talacen                             & acetaminophen/pentazocine                 \\
Vicodin HP                          & acetaminophen/hydrocodone                 \\
Vicoprofen                          & hydrocodone/ibuprofen                     \\
Percocet 7.5 / 325                  & acetaminophen/oxycodone                   \\
Lortab                              & acetaminophen/hydrocodone                 \\
Norco                               & acetaminophen/hydrocodone                 \\
Vicodin                             & acetaminophen/hydrocodone                 \\
Percocet 5 / 325                    & acetaminophen/oxycodone                   \\
Stagesic                            & acetaminophen/hydrocodone                 \\
Targiniq ER                         & naloxone/oxycodone                        \\
Xodol                               & acetaminophen/hydrocodone                 \\
Endocet                             & acetaminophen/oxycodone                   \\
Ultracet                            & acetaminophen/tramadol                    \\
Panlor SS                           & acetaminophen/caffeine/dihydrocodeine     \\
Zubsolv                             & buprenorphine/naloxone                    \\
Xartemis XR                         & acetaminophen/oxycodone                   \\
Talwin Nx                           & naloxone/pentazocine                      \\
Tylenol with Codeine                & acetaminophen/codeine                     \\
Anexsia                             & acetaminophen/hydrocodone                 \\
Darvocet-N 50                       & acetaminophen/propoxyphene                \\
Liquicet                            & acetaminophen/hydrocodone                 \\
Darvocet-N 100                      & acetaminophen/propoxyphene                \\
Trezix                              & acetaminophen/caffeine/dihydrocodeine     \\
Percodan                            & aspirin/oxycodone                         \\
Darvocet A500                       & acetaminophen/propoxyphene                \\
Percocet 2.5 / 325                  & acetaminophen/oxycodone                   \\
Balacet                             & acetaminophen/propoxyphene                \\
Aceta w/ Codeine                    & acetaminophen/codeine                     \\
Zamicet                             & acetaminophen/hydrocodone                 \\
Embeda                              & morphine/naltrexone                       \\
Bunavail                            & buprenorphine/naloxone                    \\
Tylenol with Codeine \#3            & acetaminophen/codeine                     \\
Narvox                              & acetaminophen/oxycodone                   \\
Zydone                              & acetaminophen/hydrocodone                 \\
Tylenol with Codeine \#4            & acetaminophen/codeine                     \\
Capital w/ Codeine                  & acetaminophen/codeine                     \\
Co-Gesic                            & acetaminophen/hydrocodone                 \\
Cocet Plus                          & acetaminophen/codeine                     \\
Codrix                              & acetaminophen/codeine                     \\
Dolacet                             & acetaminophen/hydrocodone                 \\
Dolagesic                           & acetaminophen/hydrocodone                 \\
Endodan                             & aspirin/oxycodone                         \\
Perloxx                             & acetaminophen/oxycodone                   \\
Phrenilin with Caffeine and Codeine & acetaminophen/butalbital/caffeine/codeine \\
Roxilox                             & acetaminophen/oxycodone                   \\
Synalgos-DC                         & aspirin/caffeine/dihydrocodeine           \\
Theracodophen Low 90                & acetaminophen/hydrocodone                 \\
Tramapap                            & acetaminophen/tramadol                    \\
Trycet                              & acetaminophen/propoxyphene                \\
Verdrocet                           & acetaminophen/hydrocodone                 \\
Zolvit                              & acetaminophen/hydrocodone                 \\
Astramorph PF                       & morphine                                  \\
Ionsys                              & fentanyl                                  \\
Lazanda                             & fentanyl                                  \\
Levo-Dromoran                       & levorphanol                               \\
Numorphan                           & oxymorphone                               \\
Onsolis                             & fentanyl                                  \\
Oxyfast                             & oxycodone                                 \\
Palladone                           & hydromorphone                             \\
Roxanol                             & morphine                                  \\
Roxanol-T                           & morphine                                  \\
Roxicodone Intensol                 & oxycodone                                 \\
Meperitab                           & meperidine                                \\
Methadone Diskets                   & methadone                                 \\
Actiq                               & fentanyl                                  \\
Fentora                             & fentanyl                                  \\
Subutex                             & buprenorphine                             \\
Demerol                             & meperidine                                \\
Dolophine                           & methadone                                 \\
Roxicodone                          & oxycodone                                 \\
Duragesic-25                        & fentanyl                                  \\
Infumorph                           & morphine                                  \\
Methadose                           & methadone                                 \\
Ultram ODT                          & tramadol                                  \\
Dilaudid                            & hydromorphone                             \\
Subsys                              & fentanyl                                  \\
MSIR                                & morphine                                  \\
OxyContin                           & oxycodone                                 \\
Paregoric                           & opium                                     \\
Duragesic-100                       & fentanyl                                  \\
Abstral                             & fentanyl                                  \\
Oxydose                             & oxycodone                                 \\
Stadol                              & butorphanol                               \\
Duragesic                           & fentanyl                                  \\
Duragesic-50                        & fentanyl                                  \\
Buprenex                            & buprenorphine                             \\
Zohydro ER                          & hydrocodone                               \\
Duragesic-75                        & fentanyl                                  \\
MS Contin                           & morphine                                  \\
Kadian                              & morphine                                  \\
Opana                               & oxymorphone                               \\
Opana ER                            & oxymorphone                               \\
Sublimaze                           & fentanyl                                  \\
Exalgo                              & hydromorphone                             \\
Opium Deodorized                    & opium                                     \\
Oxaydo                              & oxycodone                                 \\
Avinza                              & morphine                                  \\
Nucynta ER                          & tapentadol                                \\
Darvon-N                            & propoxyphene                              \\
OxyIR                               & oxycodone                                 \\
Nubain                              & nalbuphine                                \\
Dilaudid-HP                         & hydromorphone                             \\
Rybix ODT                           & tramadol                                  \\
Ultram ER                           & tramadol                                  \\
Butrans                             & buprenorphine                             \\
Darvon                              & propoxyphene                              \\
Oramorph SR                         & morphine                                  \\
Nucynta                             & tapentadol                                \\
Ultram                              & tramadol                                  \\
Duramorph                           & morphine                                  \\
Ryzolt                              & tramadol                                  \\
Talwin                              & pentazocine                               \\
Duragesic-12                        & fentanyl                                  \\
Alfenta                             & alfentanil                                \\
ConZip                              & tramadol                                  \\
Hysingla ER                         & hydrocodone                               \\
Belbuca                             & buprenorphine                             \\
Dazidox                             & oxycodone                                 \\
DepoDur                             & morphine liposomal                        \\
ETH-Oxydose                         & oxycodone                                 \\
Oxecta                              & oxycodone                                 \\
Probuphine                          & buprenorphine                             \\
RMS                                 & morphine                                  \\
Sufenta                             & sufentanil                                \\
Ultiva                              & remifentanil                              \\
\bottomrule
\end{longtable}
\end{center}

\newpage
\begin{center}
\footnotesize
\begin{longtable}{l|l}
\caption{ICD 9 and ICD 10 diagnosis codes relevant to OUD. Note that, all of these codes defined by \citet{weiss2020hospital}.}\\

\hline 
\multicolumn{1}{c|}{\textbf{ICD code}} & \multicolumn{1}{c}{\textbf{ICD Description}}                                     \\ \hline 
\endfirsthead

\multicolumn{2}{c}%
{{ \tablename\ \thetable{}: continued from previous page}} \\
\hline \multicolumn{1}{c|}{\textbf{Diagnosis code}} & \multicolumn{1}{c}{\textbf{Description}}                                     \\ \hline 
\endhead

\hline \multicolumn{2}{r}{{Continued on next page}} \\
\endfoot
\hline\hline
\multicolumn{2}{c}{\textbf{ICD 9 diagnosis codes}}                                                                              \\
\hline
304                                         & Opioid type dependence, unspecified                                                  \\
304.01                                      & Opioid type dependence, continuous                                                   \\
304.02                                      & Opioid type dependence, episodic                                                     \\
304.03                                      & Opioid type dependence, in remission                                                 \\
304.7                                       & Combinations of opioid type drug with any other drug dependence, unspecified         \\
304.71                                      & Combinations of opioid type drug with any other drug dependence, continuous          \\
304.72                                      & Combinations of opioid type drug with any other drug dependence, episodic            \\
304.73                                      & Combinations of opioid type drug with any other drug dependence, in remission        \\
305.5                                       & Opioid abuse, unspecified                                                            \\
305.51                                      & Opioid abuse, continuous                                                             \\
305.52                                      & Opioid abuse, episodic                                                               \\
305.53                                      & Opioid abuse, in remission                                                           \\
965                                         & Poisoning by opium (alkaloids), unspecified                                          \\
965.01                                      & Poisoning by heroin                                                                  \\
965.02                                      & Poisoning by methadone                                                               \\
965.09                                      & Poisoning by other opiates and related narcotics                                     \\
970.1                                       & Poisoning by opiate antagonists                                                      \\
E850.0                                      & Accidental poisoning by heroin                                                       \\
E850.1                                      & Accidental poisoning by methadone                                                    \\
E850.2                                      & Accidental poisoning by other opiates and related narcotics                          \\
E935.0                                      & Heroin causing adverse effects in therapeutic use                                    \\
E935.1                                      & Methadone causing adverse effects in therapeutic use                                 \\
E935.2                                      & Other opiates and related narcotics causing adverse effects in therapeutic use       \\
E940.1                                      & Adverse effects of opiate antagonists                                                \\
\hline
\multicolumn{2}{c}{\textbf{ICD 10 diagnosis codes}}                                                                             \\
\hline
\multicolumn{2}{l}{\textbf{Opioid abuse/dependence}}                                                                               \\
\hline
F11.10                                      & Opioid abuse, uncomplicated                                                          \\
F11.120                                     & Opioid abuse with intoxication, uncomplicated                                        \\
F11.121                                     & Opioid abuse with intoxication, delirium                                             \\
F11.122                                     & Opioid abuse with intoxication, with perceptual disturbance                          \\
F11.129                                     & Opioid abuse with intoxication, unspecified                                          \\
F11.14                                      & Opioid abuse with opioid-induced mood disorder                                       \\
F11.150                                     & Opioid abuse with opioid-induced psychotic disorder, with delusions                  \\
F11.151                                     & Opioid abuse with opioid-induced psychotic disorder, with hallucinations             \\
F11.159                                     & Opioid abuse with opioid-induced psychotic disorder, unspecified                     \\
F11.181                                     & Opioid abuse with opioid-induced sexual dysfunction                                  \\
F11.182                                     & Opioid abuse with opioid-induced sleep disorder                                      \\
F11.188                                     & Opioid abuse with other opioid-induced disorder                                      \\
F11.19                                      & Opioid abuse with unspecified opioid-induced disorder                                \\
F11.20                                      & Opioid dependence, uncomplicated                                                     \\
F11.21                                      & Opioid dependence, in remission                                                      \\
F11.220                                     & Opioid dependence with intoxication, uncomplicated                                   \\
F11.221                                     & Opioid dependence with intoxication, delirium                                        \\
F11.222                                     & Opioid dependence with intoxication, with perceptual disturbance                     \\
F11.229                                     & Opioid dependence with intoxication, unspecified                                     \\
F11.23                                      & Opioid dependence with withdrawal                                                    \\
F11.24                                      & Opioid dependence with opioid-induced mood disorder                                  \\
F11.250                                     & Opioid dependence with opioid-induced psychotic disorder, with delusions             \\
F11.251                                     & Opioid dependence with opioid-induced psychotic disorder, with hallucinations        \\
F11.259                                     & Opioid dependence with opioid-induced psychotic disorder, unspecified                \\
F11.281                                     & Opioid dependence with opioid-induced sexual dysfunction                             \\
F11.282                                     & Opioid dependence with opioid-induced sleep disorder                                 \\
F11.288                                     & Opioid dependence with other opioid-induced disorder                                 \\
F11.29                                      & Opioid dependence with unspecified opioid-induced disorder                           \\
\hline
\multicolumn{2}{l}{\textbf{Opioid use}}                                                                                             \\
\hline
F11.90                                      & Opioid use, unspecified, uncomplicated                                               \\
F11.920                                     & Opioid use, unspecified with intoxication, uncomplicated                             \\
F11.921                                     & Opioid use, unspecified with intoxication delirium                                   \\
F11.922                                     & Opioid use, unspecified with intoxication, with perceptual disturbance               \\
F11.929                                     & Opioid use, unspecified with intoxication, unspecified                               \\
F11.93                                      & Opioid use, unspecified, with withdrawal                                             \\
F11.94                                      & Opioid use, unspecified, with opioid-induced mood disorder                           \\
F11.950                                     & Opioid use, unspecified with opioid-induced psychotic disorder, with delusions       \\
F11.951                                     & Opioid use, unspecified with opioid-induced psychotic disorder, with hallucinations  \\
F11.959                                     & Opioid use, unspecified with opioid-induced psychotic disorder, unspecified          \\
F11.981                                     & Opioid use, unspecified with opioid-induced sexual dysfunction                       \\
F11.982                                     & Opioid use, unspecified with opioid-induced sleep disorder                           \\
F11.988                                     & Opioid use, unspecified with other opioid-induced disorder                           \\
F11.99                                      & Opioid use, unspecified, with unspecified opioid-induced disorder                    \\
\hline
\multicolumn{2}{l}{\textbf{Poisoning}}                                                                                             \\
\hline
T40.0X1A                                    & Poisoning by opium, accidental (unintentional), initial encounter                    \\
T40.0X1D                                    & Poisoning by opium, accidental (unintentional), subsequent encounter                 \\
T40.0X2A                                    & Poisoning by opium, intentional self-harm, initial encounter                         \\
T40.0X2D                                    & Poisoning by opium, intentional self-harm, subsequent encounter                      \\
T40.0X3A                                    & Poisoning by opium, assault, initial encounter                                       \\
T40.0X3D                                    & Poisoning by opium, assault, subsequent encounter                                    \\
T40.0X4A                                    & Poisoning by opium, undetermined, initial encounter                                  \\
T40.0X4D                                    & Poisoning by opium, undetermined, subsequent encounter                               \\
T40.1X1A                                    & Poisoning by heroin, accidental (unintentional), initial encounter                   \\
T40.1X1D                                    & Poisoning by heroin, accidental (unintentional), subsequent encounter                \\
T40.1X2A                                    & Poisoning by heroin, intentional self-harm, initial encounter                        \\
T40.1X2D                                    & Poisoning by heroin, intentional self-harm, subsequent encounter                     \\
T40.1X3A                                    & Poisoning by heroin, assault, initial encounter                                      \\
T40.1X3D                                    & Poisoning by heroin, assault, subsequent encounter                                   \\
T40.1X4A                                    & Poisoning by heroin, undetermined, initial encounter                                 \\
T40.1X4D                                    & Poisoning by heroin, undetermined, subsequent encounter                              \\
T40.2X1A                                    & Poisoning by other opioids, accidental (unintentional), initial encounter            \\
T40.2X1D                                    & Poisoning by other opioids, accidental (unintentional), subsequent encounter         \\
T40.2X2A                                    & Poisoning by other opioids, intentional self-harm, initial encounter                 \\
T40.2X2D                                    & Poisoning by other opioids, intentional self-harm, subsequent encounter              \\
T40.2X3A                                    & Poisoning by other opioids, assault, initial encounter                               \\
T40.2X3D                                    & Poisoning by other opioids, assault, subsequent encounter                            \\
T40.2X4A                                    & Poisoning by other opioids, undetermined, initial encounter                          \\
T40.2X4D                                    & Poisoning by other opioids, undetermined, subsequent encounter                       \\
T40.3X1A                                    & Poisoning by methadone, accidental (unintentional), initial encounter                \\
T40.3X1D                                    & Poisoning by methadone, accidental (unintentional), subsequent encounter             \\
T40.3X2A                                    & Poisoning by methadone, intentional self-harm, initial encounter                     \\
T40.3X2D                                    & Poisoning by methadone, intentional self-harm, subsequent encounter                  \\
T40.3X3A                                    & Poisoning by methadone, assault, initial encounter                                   \\
T40.3X3D                                    & Poisoning by methadone, assault, subsequent encounter                                \\
T40.3X4A                                    & Poisoning by methadone, undetermined, initial encounter                              \\
T40.3X4D                                    & Poisoning by methadone, undetermined, subsequent encounter                           \\
T40.4X1A                                    & Poisoning by synthetic narcotics, accidental (unintentional), initial encounter      \\
T40.4X1D                                    & Poisoning by synthetic narcotics, accidental (unintentional), subsequent encounter   \\
T40.4X2A                                    & Poisoning by other synthetic narcotics, intentional self-harm, initial encounter     \\
T40.4X2D                                    & Poisoning by other synthetic narcotics, intentional self-harm, subsequent encounter  \\
T40.4X3A                                    & Poisoning by other synthetic narcotics, assault, initial encounter                   \\
T40.4X3D                                    & Poisoning by other synthetic narcotics, assault, subsequent encounter                \\
T40.4X4A                                    & Poisoning by synthetic narcotics, undetermined, initial encounter                    \\
T40.4X4D                                    & Poisoning by synthetic narcotics, undetermined, subsequent encounter                 \\
T40.601A                                    & Poisoning by unspecified narcotics, accidental (unintentional), initial encounter    \\
T40.601D                                    & Poisoning by unspecified narcotics, accidental (unintentional), subsequent encounter \\
T40.602A                                    & Poisoning by unspecified narcotics, intentional self-harm, initial encounter         \\
T40.602D                                    & Poisoning by unspecified narcotics, intentional self-harm, subsequent encounter      \\
T40.603A                                    & Poisoning by unspecified narcotics, assault, initial encounter                       \\
T40.603D                                    & Poisoning by unspecified narcotics, assault, subsequent encounter                    \\
T40.604A                                    & Poisoning by unspecified narcotics, undetermined, initial encounter                  \\
T40.604D                                    & Poisoning by unspecified narcotics, undetermined, subsequent encounter               \\
T40.691A                                    & Poisoning by other narcotics, accidental (unintentional), initial encounter          \\
T40.691D                                    & Poisoning by other narcotics, accidental (unintentional), subsequent encounter       \\
T40.692A                                    & Poisoning by other narcotics, intentional self-harm, initial encounter               \\
T40.692D                                    & Poisoning by other narcotics, intentional self-harm, subsequent encounter            \\
T40.693A                                    & Poisoning by other narcotics, assault, initial encounter                             \\
T40.693D                                    & Poisoning by other narcotics, assault, subsequent encounter                          \\
T40.694A                                    & Poisoning by other narcotics, undetermined, initial encounter                        \\
T40.694D                                    & Poisoning by other narcotics, undetermined, subsequent encounter                     \\
\hline
\multicolumn{2}{l}{\textbf{Adverse effects}}                                                                                       \\
\hline
T40.0X5A                                    & Adverse effect of opium, initial encounter                                           \\
T40.0X5D                                    & Adverse effect of opium, subsequent encounter                                        \\
T40.2X5A                                    & Adverse effect of other opioids, initial encounter                                   \\
T40.2X5D                                    & Adverse effect of other opioids, subsequent encounter                                \\
T40.3X5A                                    & Adverse effect of methadone, initial encounter                                       \\
T40.3X5D                                    & Adverse effect of methadone, subsequent encounter                                    \\
T40.4X5A                                    & Adverse effect of synthetic narcotics, initial encounter                             \\
T40.4X5D                                    & Adverse effect of synthetic narcotic, subsequent encounter                           \\
T40.605A                                    & Adverse effect of unspecified narcotics, initial encounter                           \\
T40.605D                                    & Adverse effect of unspecified narcotics, subsequent encounter                        \\
T40.695A                                    & Adverse effect of other narcotics, initial encounter                                 \\
T40.695D                                    & Adverse effect of other narcotics, subsequent encounter                              \\
\hline
\multicolumn{2}{l}{\textbf{Long-term use of opiates}}                                                                              \\
\hline
Z79.891                                     & Long-term (current) use of opiate analgesic                                          \\
\hline
\end{longtable}
\end{center}

\newpage
\subsection{Detailed Descriptions on the Categories}
\label{apx:annotation_scheme}
\paragraph{Confirmed aberrant behavior (CAB):} This class refers to behavior that more likely lead to a catastrophic adverse events. It is defined as evidence confirming loss of control of opioid use, specifically aberrant usage of opioid medications, including:
1)	Aberrant use of opioids, such as administration/consumption in a way other than described or self-escalating doses.
2)	Evidence suggesting or proving that patient has been selling or giving away opioids to others, including family members.
3)	Use of opioids for a different indication other than the indication intended by the prescriber.
4)	Phrases suggesting current use of illicit or illicitly obtained substances or misuse of legal substances (e.g. alcohol) other than prescription opioid medications.

\paragraph{Suggested aberrant behavior (SAB):} This class refers to behavior implying patient distress related to their opioid treatment. SAB includes three kinds of behavior that suggest potential misuse of opioid.
1)	Patient attempt to get extra opioid medicine like requesting for early refill, asking for increasing dosage or reporting missing/stolen opioid medication. 
2)	Patient emotions toward opioid like request of a certain opioid medication use/change/increase. 
3)	Physician concerns.

\paragraph{Opioids:} This class refers to the mention or listing of the name(s) of the opioid medication(s) that the patient is currently prescribed or has just been newly prescribed.

\paragraph{Indication:} This class indicates that patients are using opioid under instructions, such as using opioid for pain, for treatment of opioid use disorder, etc.

\paragraph{Diagnosed Opioid Dependency:} It refers to patients have the condition of being dependent on opioids, have chronic opioid use, or is undergoing opioid titration.

\paragraph{Benzodiazepines:} This class refers co-prescribed benzodiazepines (a risk factor for accidental opioid overdose \citep{sun2017association}). In this case, the patient is simply being co-prescribed benzodiazepines (with no noted evidence for abuse).

\paragraph{Medication Change:} This class indicates that the physician makes changes to the patient’s opioid regimen during this current encounter or the patient’s opioid regimen has been changed since the patient’s last encounter with the provider writing the note.

\paragraph{Central Nervous System Related:} This is defined as CNS-related terms or terms suggesting altered sensorium, including cognitive impairment, sedation, lightheadedness, intoxication and general term suggesting altered sensorium (e.g. “altered mental status”).

\paragraph{Social Determinants of Health:} This class refers to the factors in the surroundings which impact their well-being. Our dataset captured following attributes: 
\begin{itemize}
\item Marital status (single, married ...)
\item Cohabitation status (live alone, lives with others ...)
\item Educational level (graduate degree, college degree, high-school diploma ...)
\item Socioeconomic status (retired, disabled, pension, working ...)
\item Homelessness (past, present ...)
\end{itemize}




\newpage
\section{Details on the Inter-Rator Reliability for each Category}
\label{apx:IAA_per_category}

\begin{table}[!h]
    \centering
    \
    \begin{tabular}{c|c}
\hline
       Categories  &  $\kappa$ \\
\hline\hline
         Confirmed Aberrant Behaviors & 86.94\\
\hline
Suggested Aberrant Behaviors&80.00\\
\hline
Opioids&95.73\\
\hline
Indication&82.96\\
\hline
Diagnosed Opioid Dependency&74.97 \\
\hline
Benzodiazepines&98.36 \\
\hline
Medication Change&100.00 \\
\hline
Central nervous system related&97.94 \\
\hline
Social Determinants of Health&69.95\\
\hline
    \end{tabular}
    \caption{Inter-Rator Reliability for each Category}
    \label{tab:irr_per_cate}
\end{table}

\revised{Table~\ref{tab:irr_per_cate} shows the inter-rator reliability for each category. We can see that all categories show strong reliability (Diagnosed Opioid Dependency, Social Determinant of Health) or almost perfect reliability. This means that the annotation data for each category is also of high quality.}

\section{Details on the Data Augmentation with a Large Language Model}
\label{apx:data_aug}

\begin{table}[!h]
\centering
\begin{tabular}{@{}l|c@{ }|c|c@{ }|c@{}}
\hline
\multicolumn{1}{c|}{\multirow{2}{*}{Categories}} & \multicolumn{2}{c|}{BioClinicalBERT} & \multicolumn{2}{c}{T5 Paraphrasing} \\
\cline{2-5}
\multicolumn{1}{c|}{} & AUPRC & F1 & AUPRC & F1 \\
\hline
\hline
Confirmed Aberrant Behaviors&90.52\stdintable{6.54}&78.25\stdintable{7.07}&\textbf{93.86\stdintable{4.53}}&\textbf{87.36\stdintable{6.41}}\\
\hline
Suggested Aberrant Behaviors&46.04\stdintable{16.27} & 44.37\stdintable{13.02}&\textbf{65.63\stdintable{16.02}}&\textbf{57.30\stdintable{14.65}}\\
\hline
Opioids& \textbf{99.57\stdintable{0.18}} & \textbf{98.00\stdintable{0.29}}&99.35\stdintable{0.42}&97.93\stdintable{0.27}\\
\hline
Indication&\textbf{97.86\stdintable{0.90}} & 93.55\stdintable{0.94}& 96.77\stdintable{1.34}& \textbf{95.16\stdintable{1.26}}\\
\hline
Diagnosed Opioid Dependency&90.15\stdintable{7.22} &79.24\stdintable{12.97}&\textbf{92.33\stdintable{4.80}}&\textbf{86.11\stdintable{9.44}}\\
\hline
Benzodiazepines&96.89\stdintable{1.71} & \textbf{97.15\stdintable{1.55}}&\textbf{97.28\stdintable{1.51}}&96.79\stdintable{1.10}\\
\hline
Medication Change&76.33\stdintable{4.06} & 68.61\stdintable{5.14}&\textbf{78.27\stdintable{4.79}}&\textbf{74.89\stdintable{3.07}}\\
\hline
Central Nervous System Related&98.74\stdintable{0.53}& 92.81\stdintable{2.44}&\textbf{99.16\stdintable{0.48}}&\textbf{95.41\stdintable{0.83}}\\
\hline
Social Determinants of Health& \textbf{97.39\stdintable{1.83}} & 93.79\stdintable{3.08}&96.86\stdintable{3.69}&\textbf{95.75\stdintable{1.85}}\\
\hline
\end{tabular}
\caption{Experimental results of the data augmentation with a LLM's paraphrasing.}
\label{tab:paraphrasing_full}
\end{table}

Experimental results in Table~\ref{tab:paraphrasing_full} showed that the data augmentation helps to enhance the performance of aberrant behavior detection at BioClinicalBERT + Prompt-based training environment. Especially the performance of the uncommon classes, such as diagnosed opioid dependence, suggested aberrant behaviors, diagnosed opioid dependency, increased substantially. 
However, if there is already enough data and performance is high (Opioids, Indication, Benzodiazepines, Central nervous systerm related, Social determinant of health), there is a marginal difference in performance. In addition, due to the various linguistic patterns of suggested aberrant behaviors, there is still room for performance improvement by paraphrasing alone.

\end{document}